\documentclass{article}

\usepackage[final]{corl_2026} 
\hypersetup{pdftitle={Artificial Foveated Perception for Mitigating Shortcut Learning in Robotic Foundation Models}, pdfauthor={Xiatao Sun, Yuan Zhuang, Mateo Sanchez Lopez Negrete, Matei-Victor Coldea, Chen Liang, Haoyang Zhang, Che Liu, Ziyao Zeng, Shawn Li, Qian Wang, Fei Miao, Daniel Rakita}}

\usepackage{graphicx} 
\usepackage{amsmath}
\usepackage{amsfonts}
\usepackage{natbib}
\usepackage{booktabs}
\usepackage{multirow}
\usepackage{wrapfig}
\usepackage{comment}
\usepackage{arydshln}
\title{Artificial Foveated Perception for Mitigating\\Shortcut Learning in Robotic Foundation Models}

\author{
  \begin{tabular}{c}
    Xiatao Sun$^{1\dagger}$, Yuan Zhuang$^{2}$, Mateo Sanchez Lopez Negrete$^{1}$, \\
    Matei-Victor Coldea$^{1}$, Chen Liang$^{1}$, Haoyang Zhang$^{3, 5}$, Che Liu$^{4, 5}$, \\
    Ziyao Zeng$^{1}$, Shawn Li$^{5}$, Qian Wang$^{1}$, Fei Miao$^{2}$, Daniel Rakita$^{1}$
  \end{tabular} \\
  $^{1}$Yale University \quad $^{2}$University of Connecticut \\
  $^{3}$Peking University \quad $^{4}$Imperial College London \quad $^{5}$Digients \\
  $^{\dagger}$Corresponding author: \texttt{xiatao.sun@yale.edu}
}

\begin{document}
\maketitle


\vspace{-0.3in} 
\begin{figure}[ht]
\centering
\includegraphics[width=\columnwidth]{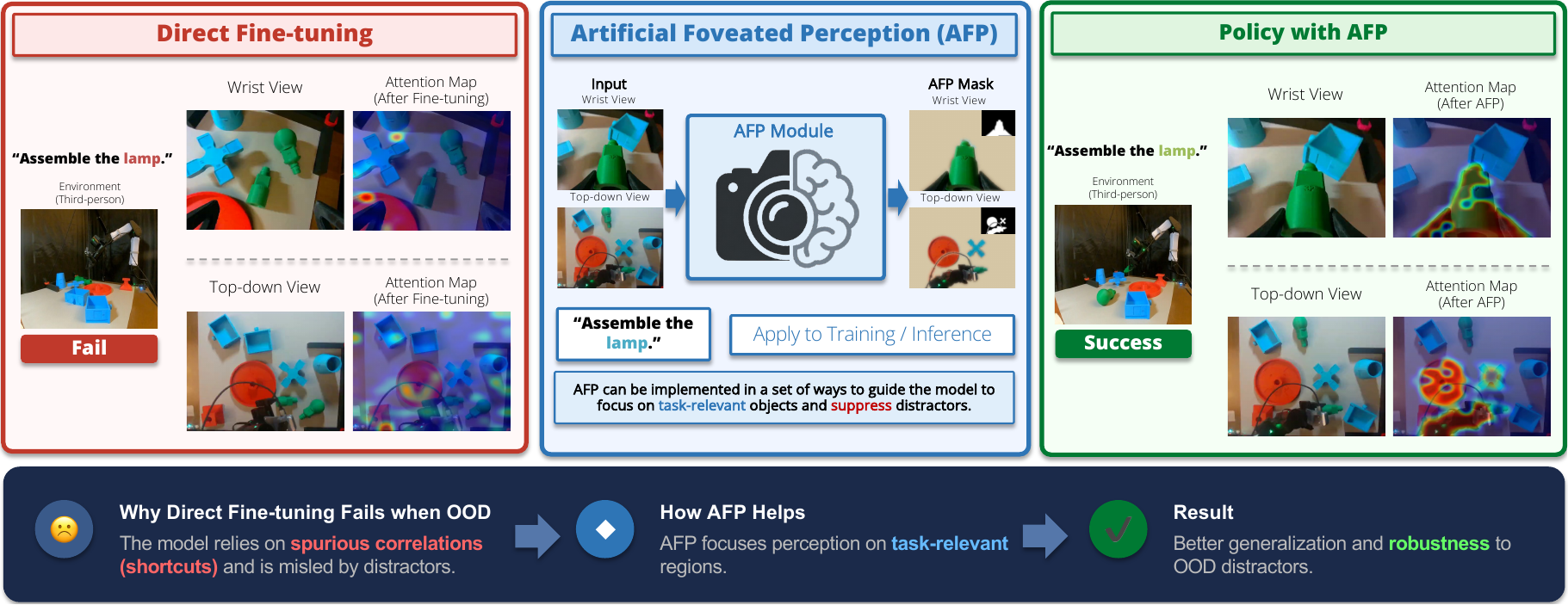}
\caption{Artificial Foveated Perception (AFP) mitigates shortcut learning in robotic foundation models by forcing or guiding their perception towards task-relevant regions while suppressing distractors. Policies from direct fine-tuning often overfit to spurious correlations and fail under out-of-distribution (OOD) perturbations, whereas AFP improves visual grounding and enables more robust policy generalization across environmental variations.}
\label{fig:teaser_img}
\vspace{-10pt}
\end{figure}


\begin{abstract}
Robotic foundation models still need task-specific fine-tuning before deployment, and the fine-tuned policies often break under modest changes in scene layout, lighting, or nearby distractors. We trace this brittleness to \textit{shortcut learning}: fine-tuning supervises actions but not the visual evidence the policy uses, so the policy can settle on scene-level correlations that predict the demonstrations without causing success. We propose Artificial Foveated Perception (AFP), a lightweight, policy-agnostic module that takes the same vision and language inputs as existing Vision-Language-Action and World Action Model pipelines and predicts task-conditioned masks over the relevant objects, the robot, and other action-critical regions. During fine-tuning the masks serve as an auxiliary grounding signal that aligns the policy's visual attention with task-relevant regions; the policy architecture is unchanged, and at inference the policy runs on the original observation stream with no AFP call in the control loop. In simulation with four robotic foundation models and on a real robot with $\pi_{0.5}$, AFP improves generalization under environmental perturbations, reduces overfitting, and shortens fine-tuning. Ablations over mask quality and grounding-loss design show that these gains come from directing policy learning toward task-relevant visual evidence. Code, data, and videos are available at \url{https://apollo-lab-yale.github.io/26-CoRL-AFP-website/}.
\end{abstract}




\section{Introduction}
\label{sec:introduction}

Robotic foundation models have made substantial progress in recent years. By leveraging representations from pretrained Vision-Language Models (VLMs) and World Models (WMs), state-of-the-art systems now demonstrate increasingly strong multi-task capability, cross-embodiment transfer, and language-conditioned control~\cite{intelligence2025pi_, kim2025openvla, shukor2025smolvla}. Despite this progress, robust deployment still often depends on task-specific adaptation. Unlike foundation models in other domains, such as Large Language Models (LLMs) with strong zero-shot inference~\cite{achiam2023gpt}, Vision-Language-Action (VLA) models and World Action Models (WAMs) typically require fine-tuning before reliable deployment in new robotic settings~\cite{zhou2025exploring, kim2025openvla, liu2025can}. However, this adaptation is frequently slow and brittle: policies may perform well in the fine-tuning environment but degrade under modest changes in object placement, lighting, camera viewpoint, background appearance, or nearby distractors~\cite{grover2025enhancing, sun2025dynamic, kachaev2025don}.

We argue that one reason for this brittleness is that standard fine-tuning provides action-level supervision without explicitly constraining which visual evidence the policy should use to adapt. Visual observations contain many cues that are predictive within a narrow fine-tuning distribution but are not causally relevant to the task. A policy may therefore exploit these cues as \textit{shortcuts}: spuriously correlated features that support low training loss in-distribution but fail to generalize under distribution shift~\cite{geirhos2020shortcut, qiu2023complexity}. In robotic manipulation, such shortcuts can arise from incidental scene context, background texture, lighting conditions, object co-occurrences, or camera-specific artifacts. A policy fine-tuned to pick up a mug, for example, should ground its actions in the mug, the robot, and their spatial relationship; however, the fine-tuning objective alone may also reward reliance on visual patterns that merely happen to correlate with successful demonstrations. Prior work in computer vision and general machine learning has shown that guidance toward task-relevant regions can help reduce shortcut learning~\cite{geirhos2020shortcut, li2025gaze, ma2023eye}; however, analogous grounding mechanisms for fine-tuning robotic foundation models remain comparatively underexplored.

In this paper, we propose Artificial Foveated Perception (AFP), a lightweight and policy-agnostic module for improving the fine-tuning of robotic foundation models. AFP takes the same vision and language inputs as the underlying VLA or WAM pipeline and predicts a task-conditioned mask over relevant objects, the robot, and other action-critical regions. We use this mask as an auxiliary grounding signal during fine-tuning: the policy is trained not only to predict actions, but also to align its visual attention with the regions identified by AFP. This encourages the policy to base its adaptation on task-relevant visual evidence rather than incidental scene-level correlations. Importantly, AFP does not require changes to the core policy architecture, and after fine-tuning the policy can run on the original observation stream without requiring AFP in the control loop.

We evaluate AFP across robotic foundation model pipelines in both simulation and real-world settings. Our experiments examine whether auxiliary foveated grounding improves fine-tuning efficiency, reduces overfitting, and improves robustness under environmental perturbations. We further analyze how mask quality and grounding-loss design affect visual grounding and policy behavior. Across these studies, AFP encourages policies to rely more strongly on task-relevant visual regions, providing a practical mechanism for making fine-tuned robotic foundation models more robust and data-efficient. 

We release the \href{https://github.com/Apollo-Lab-Yale/artificial-foveated-perception}{AFP implementation}, the \href{https://github.com/Apollo-Lab-Yale/afp-annotation-tool}{annotation tool}, and the \href{https://huggingface.co/datasets/yalesunxiatao/artificial-foveated-perception}{training dataset} to support future research on task-conditioned visual grounding in robotics; videos and additional materials are available on the \href{https://apollo-lab-yale.github.io/26-CoRL-AFP-website/}{project website}.



\section{Related Works}
\label{sec:related_works}
\subsection{Robotic Foundation Models}
Recent progress in robotic policies has been driven largely by the shift from small, task-specific models to overparameterized architectures capable of capturing complex motion behaviors and multimodal inputs~\cite{zhaolearning, chi2025diffusion}. These foundational works opened the path toward generalist robotic foundation models. Early efforts focused on Vision-Language-Action (VLA) models, which leverage pretrained VLMs to inject learned world knowledge into robotic control~\cite{kim2025openvla, tschannen2025siglip, beyer2024paligemma, barreiros2025careful}. More recently, World Action Models (WAMs) have emerged as an alternative, drawing on pretrained world model representations rather than VLMs to predict actions~\cite{bi2025motus, kim2026cosmos}. Despite architectural differences, both families share a common limitation: they require task-specific fine-tuning to perform reliably, a process that is slow and prone to overfitting~\cite{zhou2025exploring, pumacay2024colosseum}. Even after fine-tuning, these models are sensitive to minor distribution shifts---changes in scene layout, wall color, or lighting can significantly degrade policy performance.

Several lines of work address this limitation. One direction scales training data through synthetic generation or augmentation~\cite{mandlekar2023mimicgen, xuedemogen}. Another leverages the algebraic structure of neural networks to prevent catastrophic forgetting while preserving fine-tuning performance~\cite{driess2025knowledge, kim2025fine}. A third uses structured representations or improved optimization procedures to accelerate training~\cite{sun2025prism, sun2025hybrid, sun2024optimizing}. Our work takes a complementary approach: we improve generalizability and reduce fine-tuning time by directly targeting shortcut learning in the perception pipeline.

\subsection{Shortcut Learning}
Shortcut learning refers to a model's tendency to exploit spurious correlations between non-predictive input features and target outputs~\cite{maheronnaghsh2024robustness}. It has been extensively studied in computer vision but has received limited attention in robotics~\cite{steinmann2024navigating}. Proposed mitigations in computer vision span several directions.\ \citet{sagawadistributionally}\ minimize worst-group loss rather than average loss, though this requires group annotations specifying which samples are affected by each shortcut.\ \citet{arjovsky2019invariant}\ propose invariant risk minimization, penalizing predictors that generalize across some environments but not others.\ \citet{weng2024fast}\ generate shortcut-free training samples via diffusion-based image editing with self-optimized spatial masking.\ \citet{aniraj2023masking}\ demonstrate that background masking improves OOD performance---a finding that directly informs our design.

Within robotics, shortcut learning has only recently begun to receive attention.\ \citet{dai2026conla}\ address shortcut learning in the latent action encoder of action-free video, rather than in perception.\ \citet{xing2025shortcut}\ analyze and mitigate shortcut learning from a dataset structure and diversity perspective. Our work is distinct in targeting the perceptual architecture directly. To our knowledge, we are the first to mitigate shortcut learning in the perception pipeline of robotic foundation models to jointly improve generalizability and fine-tuning efficiency via masking non-salient regions.

\section{Method}
\label{sec:method}

In this section, we introduce the technical methods underlying Artificial Foveated Perception (AFP).  We describe how AFP predicts task-conditioned visual masks from RGB observations and language instructions, then show how these masks supervise the policy's visual attention through an auxiliary grounding loss.

\subsection{Overview of Artificial Foveated Perception}

AFP is designed to emulate the human fovea's ability to selectively process details in task-relevant regions while suppressing peripheral distractors. In the context of robotic foundation models, AFP acts as a dynamic visual filter. Given an RGB observation image $\mathbf{I} \in \mathbb{R}^{H \times W \times 3}$ and a natural language task description $\mathcal{T}$, AFP predicts a continuous mask $\mathbf{M} \in {[0, 1]}^{H \times W}$. This mask assigns values close to $1$ for pixels belonging to task-relevant objects and the robot's end-effector, and values close to $0$ for non-salient regions. Because AFP is lightweight and policy-agnostic, the same mask can be used across different foundation model pipelines to reduce the influence of non-salient visual evidence during fine-tuning.

\begin{figure}[t]
\centering
\includegraphics[width=\textwidth]{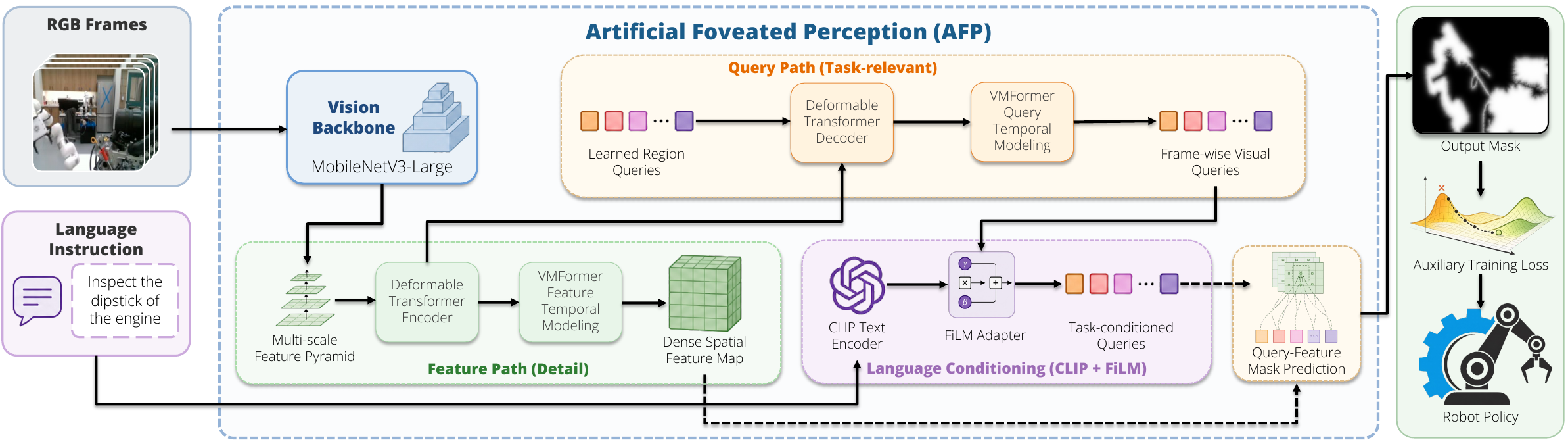}
\caption{\textbf{Architecture and Integration with Robotic Foundation Models.}
AFP predicts task-conditioned masks from RGB observations and language instructions. A feature path extracts dense multi-scale visual features, while a query path learns language-conditioned region queries that identify task-relevant objects, the robot, and other action-critical regions. The resulting mask is used during fine-tuning as an auxiliary grounding signal, aligning the policy's visual attention with task-relevant regions without modifying the core policy architecture.}
\label{fig:afp_architecture}
\vspace{-15pt}
\end{figure}

\subsection{Model Architecture}

AFP is a compact task-conditioned mask predictor. Given an observation sequence $\mathbf{I}_{1:T}$ and a task description $\mathcal{T}$, AFP predicts
$\mathbf{M}_{1:T}=f_{\mathrm{AFP}}(\mathbf{I}_{1:T},\mathcal{T})$, where each $\mathbf{M}_{t}\in[0,1]^{H\times W}$ is a dense relevance mask for frame $t$. The architecture uses two lightweight pathways: a feature path for dense spatial detail and a query path for reasoning over candidate task-relevant regions.

\paragraph{Feature and query paths.}
The feature path uses a MobileNetV3-Large backbone~\cite{howard2019searching} to extract a multi-scale feature pyramid from each frame:
\begin{equation}
    \left\{\mathbf{F}^{\ell}_{t}\right\}_{\ell=1}^{L}
    =
    \phi_{\mathrm{mv3}}(\mathbf{I}_{t}),
    \qquad
    \mathbf{F}^{\ell}_{t}
    \in
    \mathbb{R}^{H_{\ell}\times W_{\ell}\times C_{\ell}} .
\end{equation}
We use a compact CNN rather than a heavier ViT~\cite{dosovitskiyimage} or ConvNeXt~\cite{liu2022convnet} to keep AFP lightweight. Each feature level is projected to a shared hidden dimension, flattened into visual tokens, and processed by a deformable transformer encoder~\cite{zhu2020deformable}.

The query path follows the deformable DETR formulation. A small set of learned region queries $\mathbf{Q}_{0}\in\mathbb{R}^{N_{q}\times C}$ is decoded against the encoded visual memory through deformable cross-attention:
\begin{equation}
    \mathbf{Q}_{1:T}
    =
    \mathrm{Dec}_{\mathrm{def}}
    \left(
        \mathbf{Q}_{0},
        \mathrm{Enc}_{\mathrm{def}}
        \left(
            \left\{\mathbf{F}^{\ell}_{1:T}\right\}_{\ell=1}^{L}
        \right)
    \right),
\end{equation}
where $\mathrm{Enc}_{\mathrm{def}}$ and $\mathrm{Dec}_{\mathrm{def}}$ denote the deformable encoder and decoder, and $\mathbf{Q}_{1:T}$ are the frame-wise decoder query states. These queries serve as compact region descriptors for task-relevant objects and the robot end-effector. To improve temporal consistency, AFP adopts the feature-level and query-level temporal modules from VMFormer~\cite{li2024vmformer}.

\paragraph{Language conditioning and mask prediction.}
AFP conditions the decoded queries on the task description using a frozen CLIP text encoder~\cite{radford2021learning}. Given $\mathcal{T}$, CLIP produces an $\ell_{2}$-normalized text embedding $\mathbf{e}_{\mathcal{T}}$. Rather than concatenating language tokens throughout the transformer, AFP applies a lightweight FiLM adapter~\cite{perez2018film} to the decoder states:
\begin{equation}
    \widetilde{\mathbf{Q}}_{t}
    =
    \mathbf{Q}_{t}
    +
    c_{\mathcal{T}}
    \left(
        \boldsymbol{\gamma}_{\mathcal{T}}\odot\mathbf{Q}_{t}
        +
        \boldsymbol{\beta}_{\mathcal{T}}
    \right),
    \qquad
    \left(
        \boldsymbol{\gamma}_{\mathcal{T}},
        \boldsymbol{\beta}_{\mathcal{T}},
        c_{\mathcal{T}}
    \right)
    =
    g_{\psi}(\mathbf{e}_{\mathcal{T}}).
\end{equation}
Here, $g_{\psi}$ maps the text embedding to a per-channel scale $\boldsymbol{\gamma}_{\mathcal{T}}\in[-1,1]^{C}$, a per-channel shift $\boldsymbol{\beta}_{\mathcal{T}}\in\mathbb{R}^{C}$, and a scalar gate $c_{\mathcal{T}}\in[0,1]$.

Finally, a lightweight mask head fuses the encoded multi-scale features into a high-resolution feature map $\mathbf{D}_{t}\in\mathbb{R}^{C\times H^{\prime}\times W^{\prime}}$. AFP scores each image location by matching this dense feature map with the language-conditioned queries. Let $\widetilde{\mathbf{q}}_{t,i}\in\mathbb{R}^{C}$ denote the $i$-th row of $\widetilde{\mathbf{Q}}_{t}$, corresponding to the $i$-th language-conditioned region query. We compute
\begin{equation}
    s_{t,i}(\mathbf{u})
    =
    \widetilde{\mathbf{q}}_{t,i}^{\top}\mathbf{D}_{t}(\mathbf{u}),
    \qquad
    \mathbf{M}_{t}(\mathbf{u})
    =
    \sigma\left(s_{t,i^{\star}}(\mathbf{u})\right),
\end{equation}
where $\mathbf{u}$ indexes image location, $\sigma$ is the sigmoid function, and $i^{\star}$ denotes the primary saliency query. The resulting mask assigns high values to task-relevant objects, the robot end-effector, and other action-critical regions, and low values to non-salient regions.

\subsection{Model Training}

Training AFP requires task-conditioned masks that represent graded visual relevance rather than hard object segmentations. We therefore use the annotation pipeline described in Appendix~\ref{app:labeling_tool}, which allows annotators to select task-relevant objects or regions in the first frame of a trajectory and then automatically propagate the annotation through the video. The resulting masks have continuous values in $[0,1]$, capturing the robot, manipulated objects, uncertain boundaries, partial occlusions, and other action-critical regions.



Using this tool, we annotate a large collection of frames from real-world and simulated manipulation data, drawn from Open-X Embodiment~\cite{o2024open}, DROID~\cite{khazatsky2024droid}, MimicGen~\cite{mandlekar2023mimicgen}, LIBERO~\cite{liu2023libero}, and our own experimental scenes. Each training example contains an RGB observation, a language instruction, and a continuous task-relevance mask. AFP is trained on this dataset as a supervised mask predictor, learning to highlight task-relevant visual evidence while suppressing distractors. We release the \href{https://github.com/Apollo-Lab-Yale/afp-annotation-tool}{annotation tool} and the \href{https://huggingface.co/datasets/yalesunxiatao/artificial-foveated-perception}{training dataset} to support extension to new environments.

\subsection{Auxiliary Training Loss During Fine-tuning}


AFP can supervise the policy's internal visual grounding. During fine-tuning, we extract the policy attention over image tokens and average over trainable layers, heads, and query positions to obtain $\bar{\mathbf{a}}\in\mathbb{R}^{N_{\text{img}}}$, where $N_{\text{img}}$ is the number of image tokens and $\bar{a}^{(i)}$ is the policy's average attention assigned to image token $i$. We align this distribution with the pooled AFP mask $\mathbf{w}\in[0,1]^{N_{\text{img}}}$, obtained by average-pooling $\mathbf{M}$ onto the image-token grid:
\begin{equation}
    \mathcal{L}_{\text{AFP}} =
    \frac{1}{B}\sum_{b=1}^{B}
    \left\|
    \frac{\bar{\mathbf{a}}_b}{\sum_i \bar{a}^{(i)}_b}
    -
    \frac{\mathbf{w}_b}{\sum_i w^{(i)}_b}
    \right\|_2^2 .
\end{equation}
Here $B$ is the batch size, and both $\bar{\mathbf{a}}_b$ and $\mathbf{w}_b$ are normalized so the loss compares attention distributions rather than absolute attention mass. The policy is optimized with the action loss together with this grounding loss, $\mathcal{L}=\mathcal{L}_{\text{act}}+\lambda\mathcal{L}_{\text{AFP}}$, where $\lambda$ weights the AFP objective. Because the auxiliary signal should not override action learning, we treat it as a directional constraint on gradients rather than as an equally important task. Following PCGrad \cite{yu2020gradient}, let $\theta$ denote policy parameters, $\mathbf{g}_{\text{act}}=\nabla_{\theta}\mathcal{L}_{\text{act}}$ the action-loss gradient, and $\mathbf{g}_{\text{AFP}}=\nabla_{\theta}\mathcal{L}_{\text{AFP}}$ the AFP-loss gradient. When the two gradients conflict, we project the AFP gradient away from the action-gradient direction,
\begin{equation}
    \mathbf{g}_{\text{AFP}}^{\text{PC}} =
    \begin{cases}
    \mathbf{g}_{\text{AFP}} -
    \frac{\mathbf{g}_{\text{AFP}}^{\top}\mathbf{g}_{\text{act}}}
    {\|\mathbf{g}_{\text{act}}\|_2^2}
    \mathbf{g}_{\text{act}},
    & \text{if } \mathbf{g}_{\text{AFP}}^{\top}\mathbf{g}_{\text{act}} < 0, \\
    \mathbf{g}_{\text{AFP}},
    & \text{otherwise}.
    \end{cases}
\end{equation}
The update direction is then $\mathbf{g}=\mathbf{g}_{\text{act}}+\lambda\mathbf{g}_{\text{AFP}}^{\text{PC}}$. Because the AFP mask enters only as a supervision target for the policy's own attention, AFP is run on the fine-tuning data alone to produce $\mathbf{w}$, against which $\bar{\mathbf{a}}$ is compared; once training is complete, inference proceeds on the original observation stream with no AFP call in the control loop.

\section{Evaluation}
\label{sec:evaluation}

\begin{figure}[t]
    \centering
    \begin{minipage}[c]{0.49\columnwidth}
        \centering
        \includegraphics[width=\linewidth]{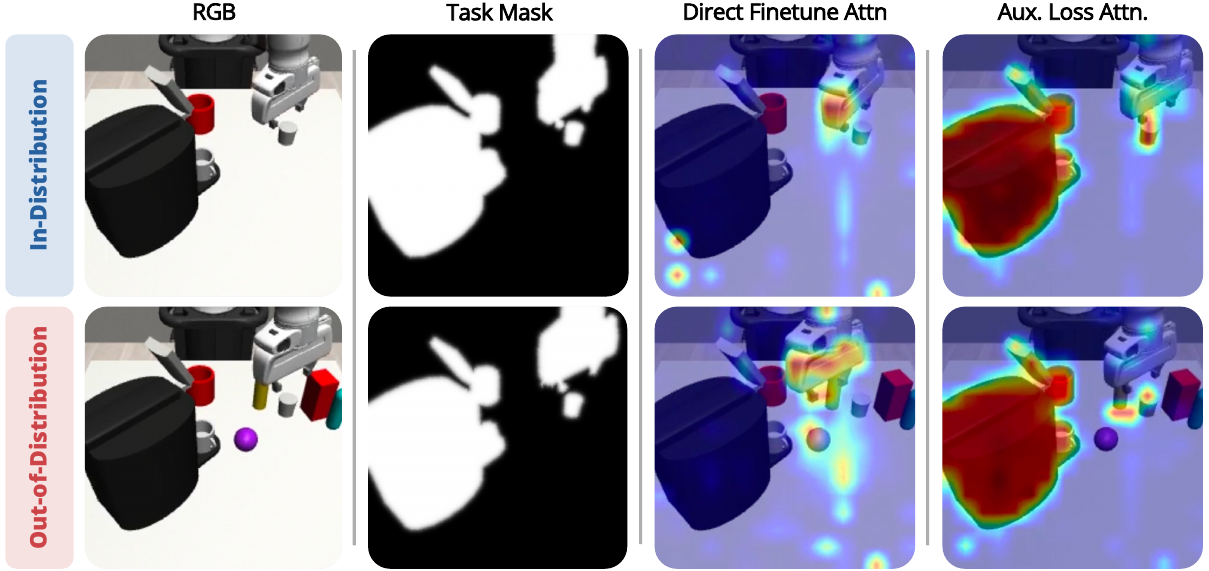}
    \end{minipage}
    \hfill
    \begin{minipage}[c]{0.49\columnwidth}
        \centering
        \resizebox{\linewidth}{!}{%
        \begin{tabular}{lcccc}
        \toprule
        \multirow{2}{*}{Metric}
        & \multicolumn{2}{c}{Direct Fine-tune}
        & \multicolumn{2}{c}{Aux. Loss} \\
        \cmidrule(lr){2-3}
        \cmidrule(lr){4-5}
        & ID & OOD
        & ID & OOD \\
        \midrule
        Soft-IoU $\uparrow$
        & 0.170 & 0.159
        & 0.934 & 0.916 \\
        EMD $\downarrow$
        & 4.389 & 4.086
        & 0.035 & 0.038 \\
        \bottomrule
        \end{tabular}%
        }
    \end{minipage}
    \caption{\textbf{Attention attribution and quantitative grounding alignment.} Left: attention attribution on the MimicGen Coffee task for direct fine-tuning and fine-tuning with the AFP auxiliary loss on in-distribution (ID) and out-of-distribution (OOD) rollout frames. Right: Soft-IoU and EMD between policy attention attribution and human-labeled task masks; higher Soft-IoU and lower EMD indicate better grounding.}
    \label{fig:strategy_compare}
    \vspace{-5pt}
\end{figure}

\subsection{Empirical Diagnosis of Shortcut Learning in Robotic Foundation Models}
\label{sec:shortcut_diagnosis}

Before evaluating policy success rates, we first examine whether shortcut learning is visible in the policy's visual grounding. Direct fine-tuning may reduce action loss while still allowing the model to rely on task-irrelevant scene cues that correlate with the demonstrations. We diagnose this behavior by comparing policy attention attribution with human-labeled task-relevance masks.

We measure alignment using Soft-IoU \cite{rahman2016optimizing} and Earth Mover's Distance (EMD) \cite{rubner2000earth}. Soft-IoU measures overlap between continuous attention maps and task masks, while EMD measures their spatial displacement. Better visual grounding corresponds to higher Soft-IoU and lower EMD.

\begin{figure}[t]
    \centering
    \includegraphics[width=\columnwidth]{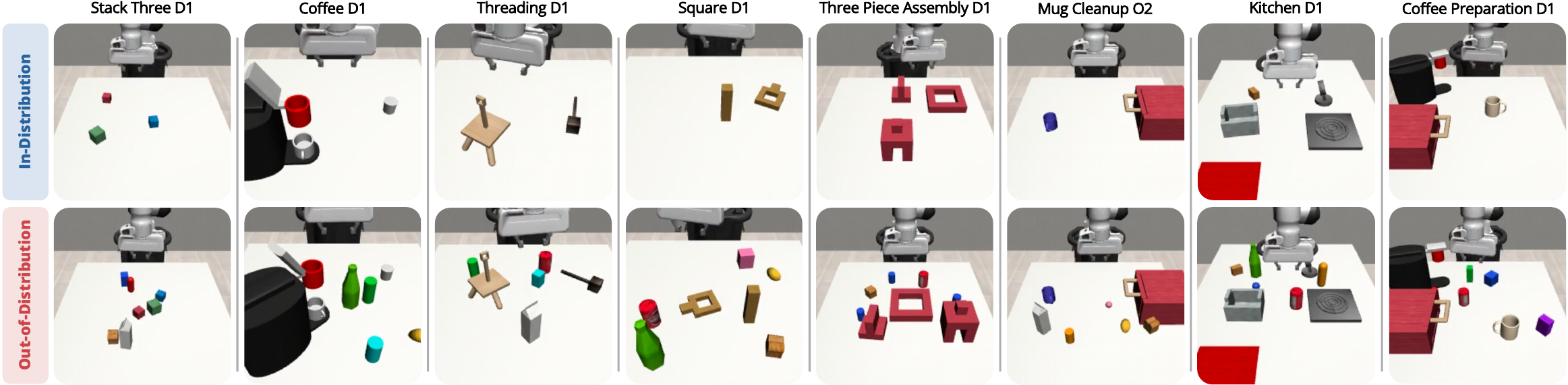}
    \caption{\textbf{ID and OOD evaluation settings.}
Top row: in-distribution scenes match the fine-tuning environments for the eight MimicGen tasks. Bottom row: out-of-distribution scenes preserve the same manipulation objective but introduce randomly placed task-irrelevant mesh objects with randomized materials, creating visual distractors that test robustness to shortcut reliance.}
    \label{fig:tasks}
    \vspace{-0pt}
\end{figure}

For each evaluation task, we generate 500 demonstration trajectories using MimicGen~\cite{mandlekar2023mimicgen} and fine-tune the policy for 20k steps. The eight MimicGen tasks used in our evaluation, together with their ID and OOD variants, are shown in Figure \ref{fig:tasks}. The ID setting matches the fine-tuning environment. To construct the OOD setting, we randomly sample 3--5 standalone mesh objects from RoboSuite \cite{zhu2020robosuite} environments that are unrelated to the current task, place them randomly on the tabletop, and assign each object a random material. This creates task-irrelevant distractors that preserve the underlying manipulation objective while perturbing the visual scene context.

We then collect 10 ID and 10 OOD rollouts for each task-model pair, yielding 20 rollout recordings per task-model pair. A human expert annotates task-relevant regions in each rollout using the tool described in Appendix \ref{app:labeling_tool}. We compute Soft-IoU and EMD for each task and report the average across all eight tasks, while keeping ID and OOD results separate. In the main paper, we report grounding results for $\pi_{0.5}$ \cite{intelligence2025pi_}; additional results on more models are provided in Appendix \ref{app:additional_shortcut_diagnosis}.

Figure \ref{fig:strategy_compare} shows that direct fine-tuning leads to poorly grounded attention. Qualitatively, attention often spreads to background or distractor regions instead of concentrating on the manipulated object and robot regions. Quantitatively, across all eight tasks for $\pi_{0.5}$, direct fine-tuning obtains Soft-IoU scores of 0.170 in ID and 0.159 in OOD, with EMD values of 4.389 and 4.086. With AFP auxiliary loss, Soft-IoU improves to 0.934 in ID and 0.916 in OOD, while EMD decreases to 0.035 and 0.038.

These results support the central motivation of AFP: direct fine-tuning can leave robotic foundation models visually misgrounded, whereas auxiliary foveated supervision explicitly aligns policy attention with task-relevant evidence.

\subsection{Simulation Experiments}


\begin{table}[t]
\centering
\caption{
\textbf{Simulation success rates across eight MimicGen tasks.}
We compare direct fine-tuning with AFP auxiliary loss on four robotic foundation models under in-distribution (ID) and out-of-distribution (OOD) settings. OOD scenes add randomized task-irrelevant mesh distractors. AFP greatly improves OOD generalization and often improves ID performance.
}
\label{tab:afp_main_results}
\resizebox{\linewidth}{!}{%
\begin{tabular}{lcccccccc}
\toprule
\multirow{2}{*}{Method}
& \multicolumn{2}{c}{Stack 3 D1}
& \multicolumn{2}{c}{Coffee D1}
& \multicolumn{2}{c}{Threading D1}
& \multicolumn{2}{c}{Square D1} \\
\cmidrule(lr){2-3}
\cmidrule(lr){4-5}
\cmidrule(lr){6-7}
\cmidrule(lr){8-9}
& ID & OOD
& ID & OOD
& ID & OOD
& ID & OOD \\
\midrule
SmolVLA
& 0.90 & 0.52
& 0.88 & 0.49
& 0.76 & 0.38
& 0.80 & 0.41 \\
SmolVLA w/ AFP
& \textbf{0.93} & \textbf{0.81}
& \textbf{0.89} & \textbf{0.78}
& \textbf{0.81} & \textbf{0.69}
& \textbf{0.85} & \textbf{0.67} \\
\hdashline[0.5pt/2pt]
OpenVLA
& 0.78 & 0.44
& 0.75 & 0.41
& 0.58 & 0.28
& 0.64 & 0.32 \\
OpenVLA w/ AFP
& \textbf{0.80} & \textbf{0.68}
& \textbf{0.76} & \textbf{0.65}
& \textbf{0.62} & \textbf{0.49}
& \textbf{0.68} & \textbf{0.55} \\
\hdashline[0.5pt/2pt]
$\pi_{0.5}$
& 0.94 & 0.61
& 0.90 & 0.58
& 0.82 & 0.46
& 0.86 & 0.49 \\
$\pi_{0.5}$ w/ AFP
& \textbf{0.96} & \textbf{0.88}
& \textbf{0.92} & \textbf{0.85}
& \textbf{0.88} & \textbf{0.77}
& \textbf{0.93} & \textbf{0.73} \\
\hdashline[0.5pt/2pt]
Motus
& 0.95 & 0.72
& 0.93 & 0.68
& 0.85 & 0.56
& 0.88 & 0.59 \\
Motus w/ AFP
& \textbf{0.95} & \textbf{0.91}
& \textbf{0.94} & \textbf{0.88}
& \textbf{0.87} & \textbf{0.82}
& \textbf{0.94} & \textbf{0.80} \\
\midrule[\heavyrulewidth]
\multirow{2}{*}{Method}
& \multicolumn{2}{c}{3 Piece Assem. D1}
& \multicolumn{2}{c}{Mug Cleanup O2}
& \multicolumn{2}{c}{Kitchen D1}
& \multicolumn{2}{c}{Coffee Prep. D1} \\
\cmidrule(lr){2-3}
\cmidrule(lr){4-5}
\cmidrule(lr){6-7}
\cmidrule(lr){8-9}
& ID & OOD
& ID & OOD
& ID & OOD
& ID & OOD \\
\midrule
SmolVLA
& 0.72 & 0.28
& 0.77 & 0.29
& 0.75 & 0.18
& 0.64 & 0.19 \\
SmolVLA w/ AFP
& \textbf{0.78} & \textbf{0.60}
& \textbf{0.79} & \textbf{0.58}
& \textbf{0.82} & \textbf{0.51}
& \textbf{0.72} & \textbf{0.48} \\
\hdashline[0.5pt/2pt]
OpenVLA
& 0.52 & 0.20
& 0.60 & 0.22
& 0.55 & 0.15
& 0.48 & 0.14 \\
OpenVLA w/ AFP
& \textbf{0.57} & \textbf{0.42}
& \textbf{0.62} & \textbf{0.46}
& \textbf{0.60} & \textbf{0.38}
& \textbf{0.53} & \textbf{0.35} \\
\hdashline[0.5pt/2pt]
$\pi_{0.5}$
& 0.79 & 0.34
& 0.79 & 0.35
& 0.82 & 0.22
& 0.70 & 0.24 \\
$\pi_{0.5}$ w/ AFP
& \textbf{0.85} & \textbf{0.68}
& \textbf{0.83} & \textbf{0.66}
& \textbf{0.89} & \textbf{0.59}
& \textbf{0.81} & \textbf{0.55} \\
\hdashline[0.5pt/2pt]
Motus
& 0.82 & 0.45
& 0.85 & 0.48
& 0.84 & 0.38
& 0.74 & 0.35 \\
Motus w/ AFP
& \textbf{0.87} & \textbf{0.74}
& \textbf{0.86} & \textbf{0.75}
& \textbf{0.88} & \textbf{0.71}
& \textbf{0.83} & \textbf{0.68} \\
\bottomrule
\end{tabular}%
}
\vspace{-5pt}
\end{table}

We next evaluate whether the improved visual grounding induced by AFP auxiliary loss translates into better policy performance. We use the same eight MimicGen tasks, demonstration data, fine-tuning budget, and ID/OOD evaluation protocol described in Section \ref{sec:shortcut_diagnosis}. In particular, each task uses 500 MimicGen-generated demonstration trajectories and 20k fine-tuning steps, and the OOD setting introduces randomly placed task-irrelevant RoboSuite mesh objects with randomized materials.

We test AFP on four robotic foundation models: SmolVLA, OpenVLA, $\pi_{0.5}$, and Motus, covering both VLA and WAM pipelines. For each model, we compare direct fine-tuning with the same model fine-tuned with AFP auxiliary loss, without changing the policy architecture or inference-time observation stream.

Table \ref{tab:afp_main_results} shows that AFP consistently improves OOD generalization across models and tasks. Direct fine-tuning often performs well in ID settings but degrades substantially under OOD perturbations, especially on longer-horizon or visually confounded tasks like Kitchen D1. AFP auxiliary loss reduces this degradation across all four models. For example, on $\pi_{0.5}$, AFP improves OOD success from 0.61 to 0.88 on Stack 3 D1, from 0.46 to 0.77 on Threading D1, and from 0.22 to 0.59 on Kitchen D1.

These gains align with the grounding diagnosis in Section \ref{sec:shortcut_diagnosis}: direct fine-tuning can leave policies visually misgrounded, while AFP encourages attention to task-relevant objects, robot regions, and contact-relevant context. This supports our hypothesis that shortcut learning is one reason robotic foundation models struggle to generalize after task-specific fine-tuning. In simulation training, we also find that the projected-gradient update described in Section \ref{sec:method} is important for stability, preventing the auxiliary grounding objective from interfering with action learning; detailed ablations are provided in Appendix \ref{app:aux_loss_stability}.

AFP also improves performance in many ID settings, indicating that the auxiliary loss is not merely an OOD regularizer. As shown in Appendix \ref{app:training_efficiency}, by directing optimization toward action-relevant visual evidence, AFP can improve fine-tuning efficiency, especially on longer-horizon tasks. 



\subsection{Real-World Experiments}

\begin{figure}[t]
    \centering
    \includegraphics[width=\columnwidth]{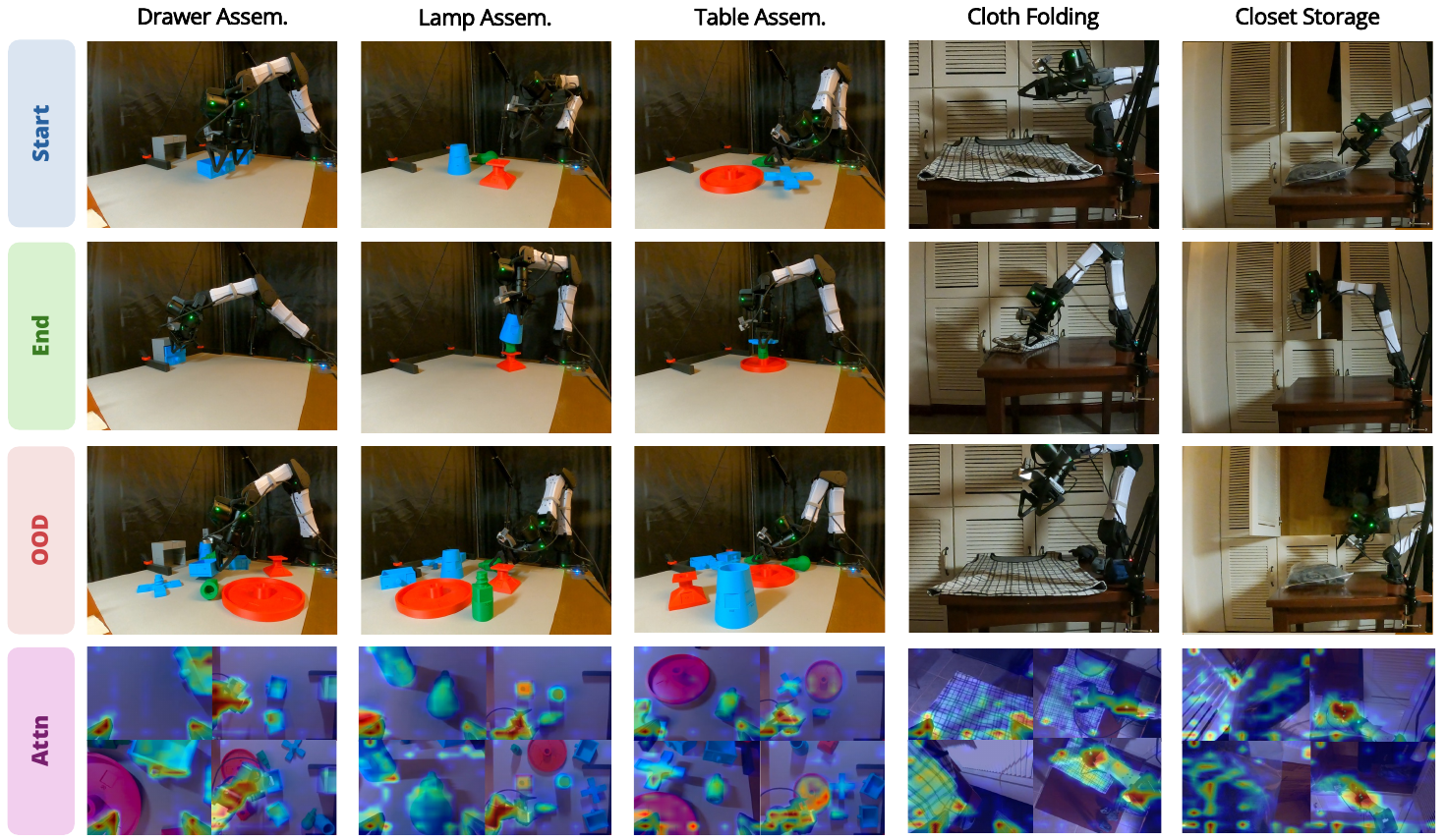}
    \caption{\textbf{Real-world ID/OOD tasks and attention attribution.}
We evaluate $\pi_{0.5}$ with AFP auxiliary loss on five physical manipulation tasks using an i2RT YAM Arm. Rows show start states, end states, OOD perturbations, and attention attribution. OOD settings introduce task-irrelevant distractors, including unused furniture parts, unseen socks, or an opened cabinet door. Each attention cell shows ID wrist, ID top, OOD wrist, and OOD top views.}
    \label{fig:real_world_tasks}
    \vspace{-5pt}
\end{figure}


\begin{table}[t]
\centering
\caption{
\textbf{Real-world success rates.}
We compare direct fine-tuning and AFP auxiliary loss for $\pi_{0.5}$ across five physical manipulation tasks. Each task uses 300 demonstration episodes and 50k fine-tuning steps, and each method is evaluated with 30 rollouts under both ID and OOD settings. AFP improves success rates on every task, with especially large gains under OOD perturbations.
}
\label{tab:afp_real_world_results}
\resizebox{\linewidth}{!}{
\begin{tabular}{lcccccccccc}
\toprule
\multirow{2}{*}{Method}
& \multicolumn{2}{c}{Drawer Assem.}
& \multicolumn{2}{c}{Lamp Assem.}
& \multicolumn{2}{c}{Table Assem.}
& \multicolumn{2}{c}{Cloth Folding}
& \multicolumn{2}{c}{Closet Storage} \\
\cmidrule(lr){2-3}
\cmidrule(lr){4-5}
\cmidrule(lr){6-7}
\cmidrule(lr){8-9}
\cmidrule(lr){10-11}
& ID & OOD
& ID & OOD
& ID & OOD
& ID & OOD
& ID & OOD \\
\midrule
$\pi_{0.5}$
& 0.73 & 0.30
& 0.57 & 0.20
& 0.40 & 0.13
& 0.67 & 0.47
& 0.63 & 0.23 \\
$\pi_{0.5}$ w/ AFP
& \textbf{0.83} & \textbf{0.67}
& \textbf{0.67} & \textbf{0.53}
& \textbf{0.50} & \textbf{0.37}
& \textbf{0.73} & \textbf{0.57}
& \textbf{0.70} & \textbf{0.53} \\
\bottomrule
\end{tabular}
}
\vspace{-10pt}
\end{table}

We further evaluate whether AFP auxiliary loss transfers to physical robot deployment. While simulation allows controlled OOD perturbations and broad model comparisons, real-world scenes introduce additional challenges such as sensor noise, lighting variation, imperfect actuation, background clutter, and more complex object appearances. We therefore test AFP on five real-world manipulation tasks using an i2RT YAM Arm, as shown in Figure \ref{fig:real_world_tasks}.

We focus on $\pi_{0.5}$ in the real-world experiments because it provides the most complete deployment support among the robotic foundation models considered in this work. For each task, we collect 300 demonstration episodes and fine-tune the policy for 50k steps. We compare direct fine-tuning against the same policy fine-tuned with AFP auxiliary loss, keeping the policy architecture and inference-time observation stream unchanged. Each task, method, and ID/OOD setting is evaluated with 30 physical rollouts, and we report success rates in Table \ref{tab:afp_real_world_results}.

The real-world tasks include three assembly tasks from FurnitureBench: Drawer Assembly, Lamp Assembly, and Table Assembly \cite{heo2025furniturebench}. For these tasks, the OOD distractors are unused 3D-printed furniture parts from the same task family, placed in the workspace but not needed for the current assembly. For Cloth Folding, the OOD distractors are socks placed on the table, which do not appear in the training set. For Closet Storage, the robot must place a folded garment into a cabinet and close the door; the OOD distractor is the right cabinet door, which remains closed throughout training but is opened during OOD evaluation. These perturbations preserve the intended task objective while introducing visually salient, task-irrelevant cues that can induce shortcut reliance.

Figure \ref{fig:real_world_tasks} qualitatively shows that AFP-trained policies retain task-relevant visual grounding in real-world scenes. The attention maps are noisier than in simulation, as expected given the higher visual complexity of physical environments, but the policy still concentrates most attribution on the manipulated objects, robot-object interaction regions, and task-relevant workspace structure. This suggests that AFP does not merely overfit to clean simulator masks; instead, the auxiliary grounding signal can preserve causal visual attention under real sensor observations.

The quantitative results confirm this trend. AFP auxiliary loss improves success rates on every real-world task in both ID and OOD settings. The gains are especially pronounced under OOD perturbations: for example, success improves from 0.30 to 0.67 on Drawer Assembly, and from 0.20 to 0.53 on Lamp Assembly. These improvements indicate that the grounding benefits observed in simulation translate to physical robot control. Overall, the real-world experiments support AFP auxiliary loss as a practical mechanism for improving policy robustness.
\section{Discussion}
\label{sec:discussion}

This work identifies shortcut learning as a visual grounding failure in robotic foundation models: direct fine-tuning can reduce action loss while leaving policy attention misaligned with task-relevant objects and robot interaction regions. AFP addresses this by training a lightweight task-conditioned mask predictor and using its continuous masks as auxiliary attention supervision during fine-tuning. Across simulation and real-world experiments, this improves grounding, OOD robustness, and often fine-tuning efficiency, offering a practical step toward grounded intelligence by anchoring robotic policies to task-relevant visual evidence, without changing the policy architecture or requiring AFP at inference time.

\subsection{Limitations}
\label{sec:limitations}
The main limitation of AFP is its reliance on task-relevance annotation. Our method first uses human-labeled structured perception data to train AFP, then uses AFP to supervise robotic foundation models. This makes the supervision explicit and reusable, but introduces annotation cost and possible bias. AFP may fail when task relevance is ambiguous, heavily occluded, or outside the mask dataset distribution. Future work can address this issue in two ways: scaling the dataset with our \href{https://github.com/Apollo-Lab-Yale/afp-annotation-tool}{annotation tool} to train stronger AFP models, or developing self-supervised grounding objectives that infer task relevance from interaction, temporal consistency, counterfactual scene changes, or policy failures.

\clearpage

\acknowledgments{We thank Ravi Khanna for help with data annotation. This work was supported by Office of Naval Research award N00014-24-1-2124.}


\bibliography{references}  

\newpage
\appendix
\section{Auxiliary Loss Projection for Stability}
\label{app:aux_loss_stability}

\begin{figure}[htbp]
    \centering
    \begin{minipage}[t]{0.49\columnwidth}
        \centering
        \includegraphics[width=\linewidth]{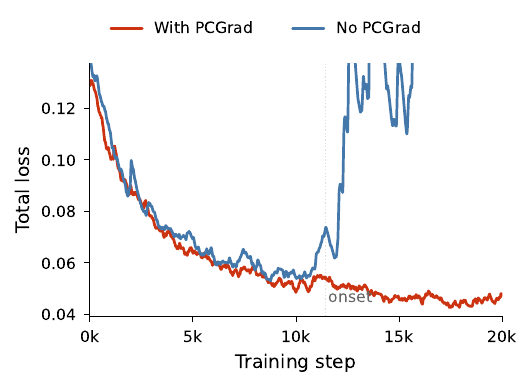}
    \end{minipage}
    \hfill
    \begin{minipage}[t]{0.49\columnwidth}
        \centering
        \includegraphics[width=\linewidth]{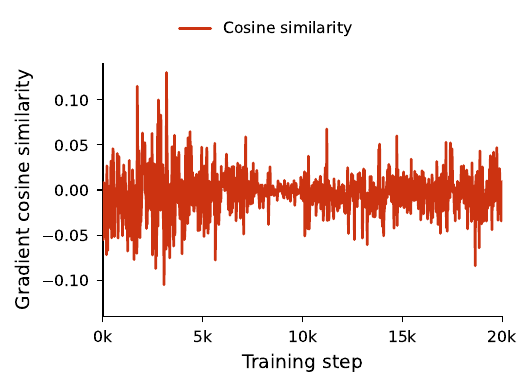}
    \end{minipage}
    \caption{
        \textbf{Projected gradients stabilize AFP fine-tuning.}
        Left: the y-axis is zoomed in to enable a reasonable comparison between the two curves; the truncated portions of the No PCGrad curve extend beyond the plotted range after the onset point. Without PCGrad-style projection, the total loss spikes after the onset point; with projection, training remains stable. Right: cosine similarity between the action-loss and AFP-loss gradients, showing frequent gradient conflict during training. Projection is triggered 11,152 times over 20k steps.}
    \label{fig:afp_loss_diagnostics}
\end{figure}

AFP auxiliary loss provides useful grounding supervision, but it also introduces an additional optimization objective that may conflict with action learning. This issue is especially important during the early stages of fine-tuning, when the policy attention is poorly grounded and the auxiliary gradient can point in a direction that reduces attention error but interferes with the action-loss update. We therefore use the projected-gradient update in Section \ref{sec:method} to treat AFP as a directional grounding constraint rather than an equally weighted task objective.

Figure \ref{fig:afp_loss_diagnostics} illustrates this effect. Without gradient projection, the total loss initially decreases but becomes unstable after the onset point, producing a sharp spike and failing to maintain stable optimization. In contrast, the PCGrad-style projected update yields a smooth and monotonic decrease throughout training. This suggests that the instability is not caused by the AFP signal itself, but by unresolved conflict between the action-loss and auxiliary-loss gradients.

The gradient cosine similarity provides further evidence for this interpretation. Early in training, the cosine similarity between the action gradient and the AFP gradient fluctuates substantially and often becomes negative, indicating frequent objective conflict. As training progresses, the similarity concentrates closer to zero, suggesting that the grounding objective becomes increasingly compatible with action learning. Over 20k training steps, the projection is triggered 11,152 times, showing that gradient conflict is common rather than incidental. By removing only the conflicting component of the AFP gradient, the projected update preserves the grounding benefit of AFP while preventing the auxiliary loss from destabilizing policy fine-tuning.

\section{Labeling Tool}
\label{app:labeling_tool}

\begin{figure}[htbp]
\centering
\includegraphics[width=0.6\columnwidth]{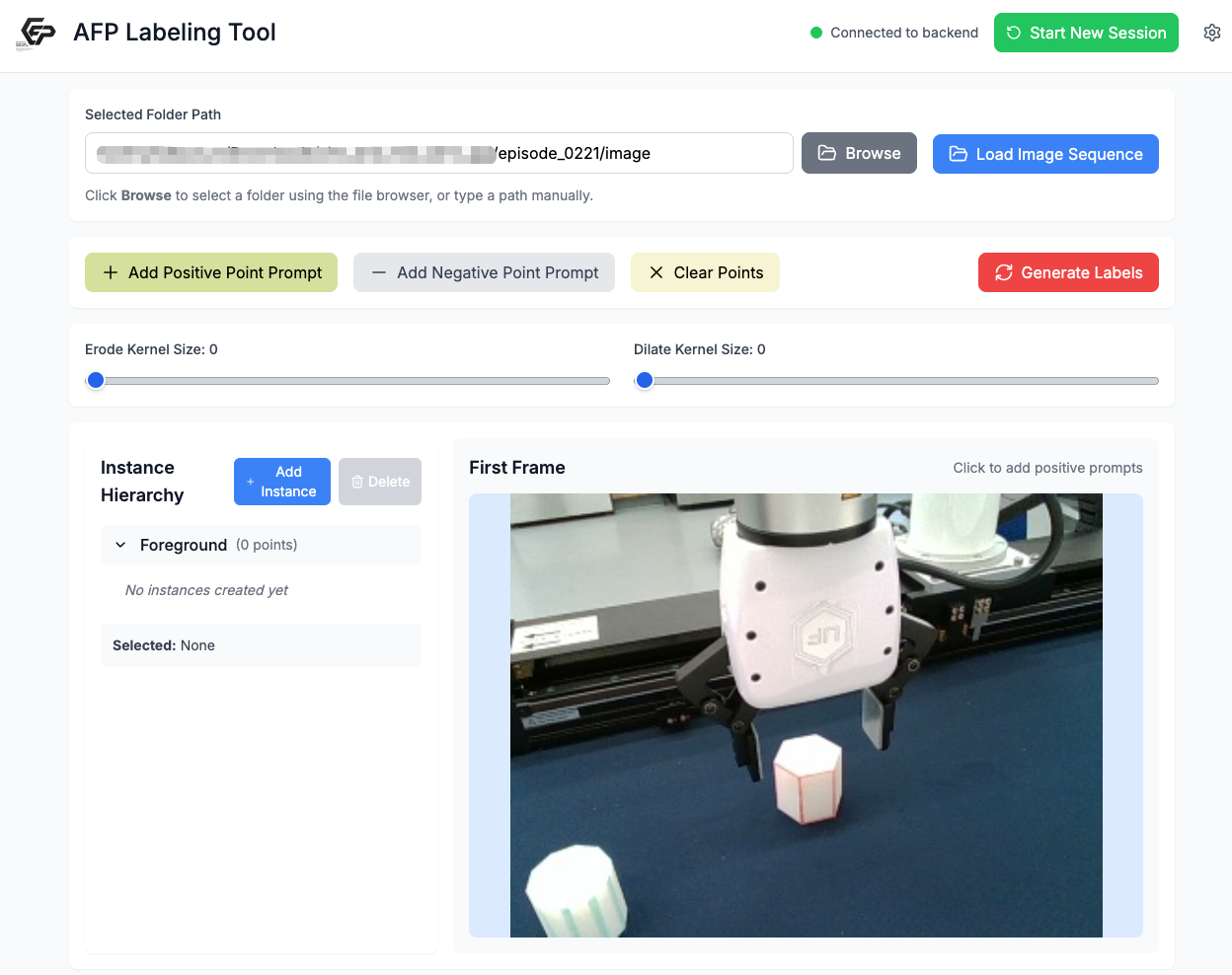}
\caption{\textbf{AFP labeling tool.}
The interface allows annotators to load a local image sequence, provide positive and negative point prompts on a key frame, manage multiple task-relevant instances, and generate continuous alpha masks for the full trajectory.}
\label{fig:afp_labeling_tool}
\end{figure}

AFP requires task-conditioned masks that behave like continuous relevance fields rather than hard binary segmentations. Standard annotation tools are typically designed for object segmentation, where each pixel is assigned to foreground or background. This is poorly matched to foveated perception and attention supervision: task relevance can be graded around object boundaries, contact regions, partially occluded objects, and robot interaction areas. We therefore develop the AFP Labeling Tool to efficiently produce continuous masks for robot manipulation trajectories.

The tool is a local, browser-based annotation system for image sequences. An annotator selects an episode folder, adds positive and negative point prompts on a key frame, and optionally separates different semantic regions using multiple instance layers. The point prompts are passed to SAM 2 \cite{ravi2025sam} to obtain an interactive key-frame mask. The selected instance masks are then merged and used to initialize MatAnyone \cite{yang2025matanyone}, which propagates the annotation through the full sequence and outputs per-frame foreground images and alpha masks.

This design substantially reduces the cost of labeling long robot trajectories: the annotator only needs to specify task-relevant regions on a small number of key frames, while temporal propagation produces dense labels for the full episode. The continuous alpha output is especially important for AFP because it provides a soft supervision target that better matches the policy attention distribution used in our auxiliary loss. 
The tool\footnote{\url{https://github.com/Apollo-Lab-Yale/afp-annotation-tool}} is released together with AFP to support future work on task-conditioned visual grounding and to allow the robotics community to extend AFP to new tasks, embodiments, and environments.

\section{Additional Empirical Findings on Shortcut Learning}
\label{app:additional_shortcut_diagnosis}

\begin{figure}[htbp]
    \centering
    \begin{minipage}[c]{0.49\columnwidth}
        \centering
        \includegraphics[width=\linewidth]{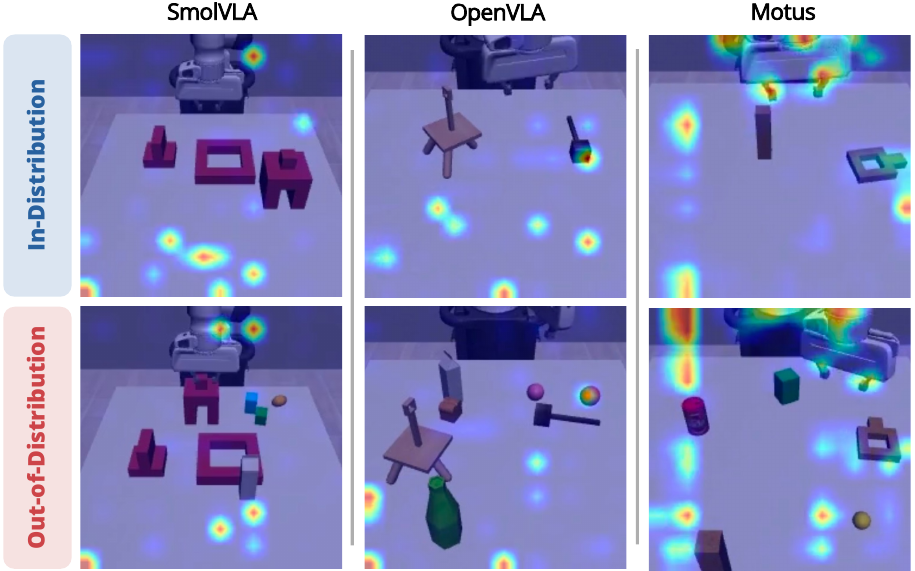}
    \end{minipage}
    \hfill
    \begin{minipage}[c]{0.49\columnwidth}
        \centering
        \small
        \setlength{\tabcolsep}{3.6pt}
        \renewcommand{\arraystretch}{1.08}
        \resizebox{\linewidth}{!}{%
        \begin{tabular}{llcccc}
        \toprule
        \multirow{2}{*}{Backbone} & \multirow{2}{*}{Method}
        & \multicolumn{2}{c}{Soft-IoU $\uparrow$}
        & \multicolumn{2}{c}{EMD $\downarrow$} \\
        \cmidrule(lr){3-4}\cmidrule(lr){5-6}
        & & ID & OOD & ID & OOD \\
        \midrule
        \multirow{2}{*}{SmolVLA}
        & Direct & 0.124 & 0.105 & 4.647 & 5.322 \\
        & Aux. Loss & \textbf{0.795} & \textbf{0.763} & \textbf{0.075} & \textbf{0.081} \\
        \midrule
        \multirow{2}{*}{OpenVLA}
        & Direct & 0.157 & 0.133 & 4.601 & 4.883 \\
        & Aux. Loss & \textbf{0.880} & \textbf{0.849} & \textbf{0.063} & \textbf{0.067} \\
        \midrule
        \multirow{2}{*}{Motus}
        & Direct & 0.176 & 0.173 & 4.433 & 4.802 \\
        & Aux. Loss & \textbf{0.895} & \textbf{0.861} & \textbf{0.031} & \textbf{0.035} \\
        \bottomrule
        \end{tabular}%
        }
    \end{minipage}
    \caption{
\textbf{Shortcut learning appears across robotic foundation models.}
Left: qualitative attention attribution for SmolVLA, OpenVLA, and Motus on MimicGen scenarios different from the main-paper Coffee example. Direct fine-tuning often attends to distractors or diffuse background regions in both ID and OOD settings. Right: grounding alignment between policy attention attribution and expert task masks. Across all backbones, AFP auxiliary loss substantially improves Soft-IoU and reduces EMD, indicating more consistent task-relevant visual grounding.
}
\label{fig:additional_empirical_findings}
\end{figure}

The main paper reports grounding diagnosis results for $\pi_{0.5}$. Here we provide additional evidence that shortcut learning is not specific to a single robotic foundation model. We repeat the same attention--mask alignment analysis for SmolVLA, OpenVLA, and Motus, using the ID/OOD rollout protocol described in Section \ref{sec:shortcut_diagnosis}. To further demonstrate that the phenomenon is not tied to the Coffee scenario used in the main qualitative example, Figure \ref{fig:additional_empirical_findings} visualizes attention maps from different MimicGen scenes.

Across all three backbones, direct fine-tuning produces poorly grounded attention. Qualitatively, the attention maps often assign mass to background structure, distractor objects, or spatially diffuse regions rather than the task-relevant object and robot interaction area. Quantitatively, direct fine-tuning yields low Soft-IoU and high EMD under both ID and OOD settings. This indicates that visual misgrounding persists even when the policy family changes from compact VLA models to larger VLA or WAM-style backbones.

AFP auxiliary loss consistently improves grounding alignment. For SmolVLA, Soft-IoU improves from 0.124 to 0.795 in ID and from 0.105 to 0.763 in OOD. OpenVLA and Motus show the same pattern, with large increases in Soft-IoU and near-zero EMD after auxiliary grounding supervision. These results support the central claim of the paper: shortcut learning is a common failure mode across robotic foundation models, and AFP mitigates it by explicitly aligning policy attention with task-relevant visual evidence.

\section{Training Efficiency}
\label{app:training_efficiency}

\begin{figure}[htbp]
    \centering
    \includegraphics[width=\columnwidth]{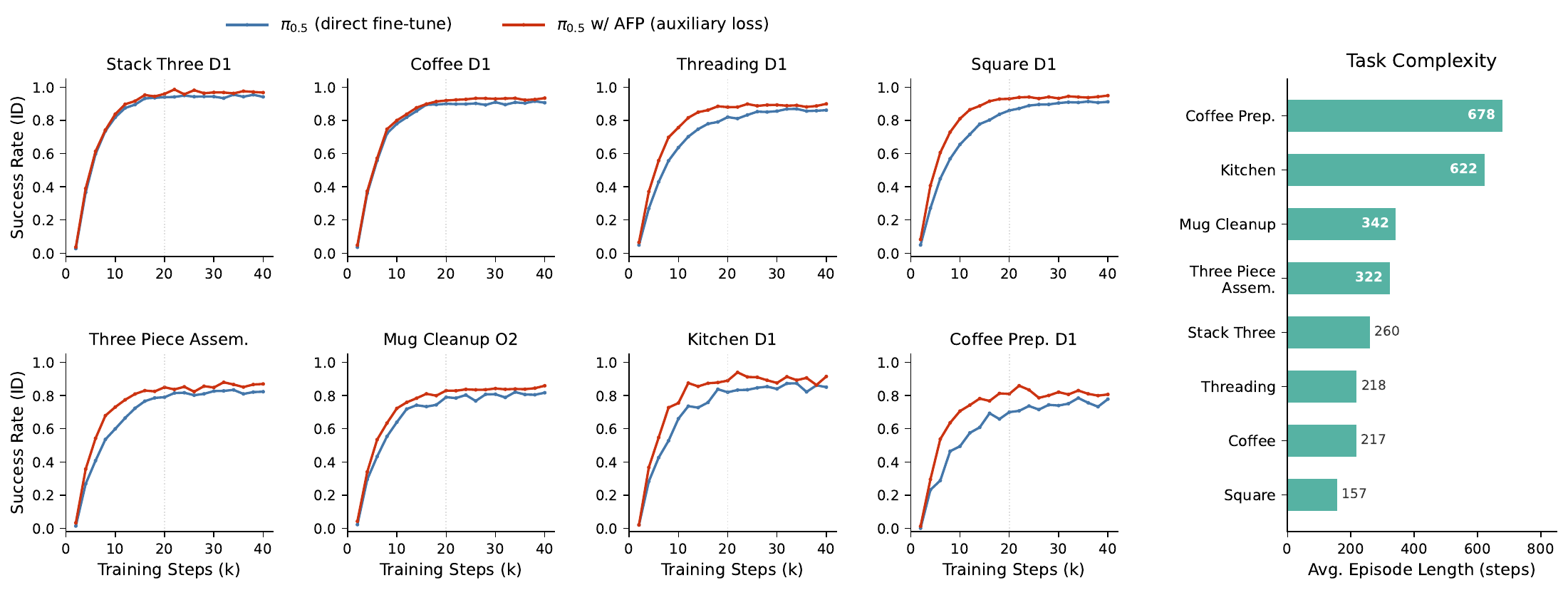}
    \caption{
    \textbf{AFP improves convergence speed, especially on longer-horizon tasks.}
    Left: ID success rate during fine-tuning for $\pi_{0.5}$ with and without AFP auxiliary loss across eight MimicGen tasks. Right: task complexity measured by average episode length. AFP provides the largest convergence gains on longer or more visually complex tasks, where task-relevant evidence is sparse within the full observation stream.
    }
    \label{fig:convergence_speed}
\end{figure}

In addition to improving OOD robustness, AFP auxiliary loss can improve fine-tuning efficiency by directing the policy toward action-relevant visual evidence. Figure \ref{fig:convergence_speed} compares the ID success rate of $\pi_{0.5}$ under direct fine-tuning and AFP auxiliary loss across the eight MimicGen tasks. Both methods eventually reach similar performance on simple short-horizon tasks such as Square D1 and Coffee D1, where the task-relevant object is visually salient and the behavior can be learned with relatively few steps. In these cases, AFP provides only a modest convergence benefit.

The advantage becomes more pronounced on longer-horizon or visually more complex tasks. On tasks such as Kitchen D1, Coffee Preparation D1, and Three Piece Assembly D1, AFP reaches high success substantially earlier than direct fine-tuning. These tasks require the policy to maintain attention over multiple objects, interaction phases, or spatially separated subgoals. The task-complexity plot on the right of Figure \ref{fig:convergence_speed} shows that these tasks also have longer average episode lengths. This suggests that the benefit of AFP grows when the policy must learn from longer trajectories with more opportunities to attend to incidental scene cues.

We attribute this improvement to the auxiliary grounding signal provided by AFP. By highlighting task-relevant objects, robot regions, and interaction context during fine-tuning, AFP reduces the burden on the policy to discover the relevant visual structure from action supervision alone. This produces better sample efficiency, especially in tasks where the relevant causal evidence is sparse relative to the full observation stream.

\end{document}